\documentclass{article}
\usepackage{ijcai26}

\usepackage{inconsolata}
\usepackage{amssymb} 
\usepackage{xcolor}  

\usepackage{times}
\usepackage{soul}
\usepackage{url}
\usepackage[hidelinks]{hyperref}
\usepackage[utf8]{inputenc}
\usepackage[small]{caption}
\usepackage{graphicx}
\usepackage{amsmath}
\usepackage{amsthm}
\usepackage{booktabs}
\usepackage{algorithm}
\usepackage{algorithmic}
\usepackage[switch]{lineno}
\usepackage{multirow}
\usepackage{multicol} 
\usepackage[T1]{fontenc}

\usepackage{booktabs}   
 
\usepackage{enumitem}    
\setlist[itemize]{leftmargin=1.5em, itemindent=0pt, labelindent=0pt}
\usepackage{makecell}
\usepackage{array}
\usepackage{tikz}
\usepackage{colortbl}
\usepackage{tcolorbox}
\tcbuselibrary{breakable} 

\usepackage{listings}      
\usepackage{ifthen}         
\usepackage{xparse}         
\usepackage{calc}           
\usepackage{bm}

\newcommand{\tool}{\textsc{RISE}} 

\title{Emerging from Ground: Addressing Intent Deviation in Tool-Using Agents via Deriving Real Calls into Virtual Trajectories}

\author{
Qian Xiong$^{1,2,3,4}$\thanks{Equal contribution, co-first author.}\and
Yuekai Huang$^{2,3,4}$\footnotemark[1]\and
Bo Yang$^1$\and
Yujia Zheng$^5$\and
Tianhao Li$^5$\and \\
Ziyou Jiang$^{2,3,4}$\and
Zhiyuan Chang$^{2,3,4}$\and
Zhaoyang Li$^{2,3,4}$\and
Huanxiang Feng$^{2,3,4}$\and
Mingyang Li$^{2,3,4}$\thanks{Corresponding author.}\\
\affiliations
$^1$Beijing Forestry University\\ 
$^2$State Key Laboratory of Complex System Modeling and Simulation Technology, Beijing, China\\
$^3$Institute of Software, Chinese Academy of Sciences, Beijing, China\\
$^4$University of Chinese Academy of Sciences, Beijing, China\\
$^5$Duke University\\
\emails
\texttt{\normalsize \{x\_qianq, yangbo\}@bjfu.edu.cn}\\ 
\texttt{\normalsize \{yuekai2018,ziyou2019,zhiyuan2019,lizhaoyang2024,mingyang2017\}@iscas.ac.cn}\\ 
\texttt{\normalsize \{yujia.zheng, tianhao.li\}@duke.edu}, 
\texttt{\normalsize fenghuanxiang25@mails.ucas.ac.cn}\\
}

\begin{document}

\maketitle

\begin{abstract}
LLMs have advanced tool-using agents for real-world applications, yet they often lead to unexpected behaviors or results. Beyond obvious failures, the subtle issue of ``intent deviation'' severely hinders reliable evaluation and performance improvement.
Existing post-training methods generally leverage either real system samples or virtual data simulated by LLMs.
However, the former is costly due to reliance on hand-crafted user requests, while the latter suffers from distribution shift from the real tools in the wild. Additionally, both methods lack negative samples tailored to intent deviation scenarios, hindering effective guidance on preference learning.
We introduce {\tool}, a ``Real-to-Virtual’’ method designed to mitigate intent deviation. 
Anchoring on verified tool primitives, {\tool} synthesizes virtual trajectories and generates diverse negative samples through mutation on critical parameters. 
With synthetic data, {\tool} fine-tunes backbone LLMs via the two-stage training for intent alignment.
Evaluation results demonstrate that data synthesized by {\tool} achieve promising results in eight metrics covering user requires, execution trajectories and agent responses.
Integrating with training, {\tool} achieves an average 35.28\% improvement in $Acc_{\text{task}}$ (task completion) and 23.27\% in $Acc_{\text{intent}}$ (intent alignment), outperforming SOTA baselines by 1.20--42.09\% and 1.17--54.93\% respectively.
\end{abstract}

\section{Introduction}
\label{sec:introduction}
LLMs have driven the evolution of tool-using agents by extending text generation with external tool invocation~\cite{luo2025large,mohammadi2025evaluation}, grounding language understanding in executable actions, and enabling impactful real-world applications~\cite{zhang2025survey,zhu2025large,tang2025llm,ge2025survey}.
However, due to the inherent limitations, tool-using agents can produce many unexpected behaviors or results. 
A typical example involves parameter hallucination, such as incompatible names or values in parameter filling, resulting in invocation failures and task disruption~\cite{xiong2025butterfly,patil2024gorilla}.

In addition to obvious errors, there exists a more nuanced challenge known as \textbf{Intent Deviation}. 
This characterizes a state where an agent appears functional, yet its trajectories deviate from the user’s actual intent. 
Concretely, this is evidenced by inappropriate tool selection during the planning (e.g., invoking \textit{Baidu} instead of the expected \textit{Bing} for news search) or unexpected parameter entry during execution (e.g., parameter value fills \textit{January 20th} instead of the requested \textit{January 19th}).
Unlike failures with explicit traces, the covert nature of the intent deviation presents distinct difficult for identification and assessment.

A common approach to address such issues involves curating specialized datasets, 
integrated with post-training techniques for capability enhancement~\cite{qin2023toolllmfacilitatinglargelanguage,chen2024towards,wang2025toolgenunifiedtoolretrieval,chen2025advancingtoolaugmentedlargelanguage}.
In this framework, constructing high-quality datasets aligned with predefined objectives remains a major challenge.
One common approach relies on the trajectory sampling from real agent systems
~\cite{huang2024metatoolbenchmarklargelanguage,qin2023toolllmfacilitatinglargelanguage,wang2024llmsimaginariumtoollearning}. 
While it ensures authenticity, the request-driven nature incurs high costs in input diversity, tool coverage, and ground truth labeling (needing to carefully analyze the execution trajectory).

\begin{figure}[t]
    \centering
    \includegraphics[width=\linewidth]{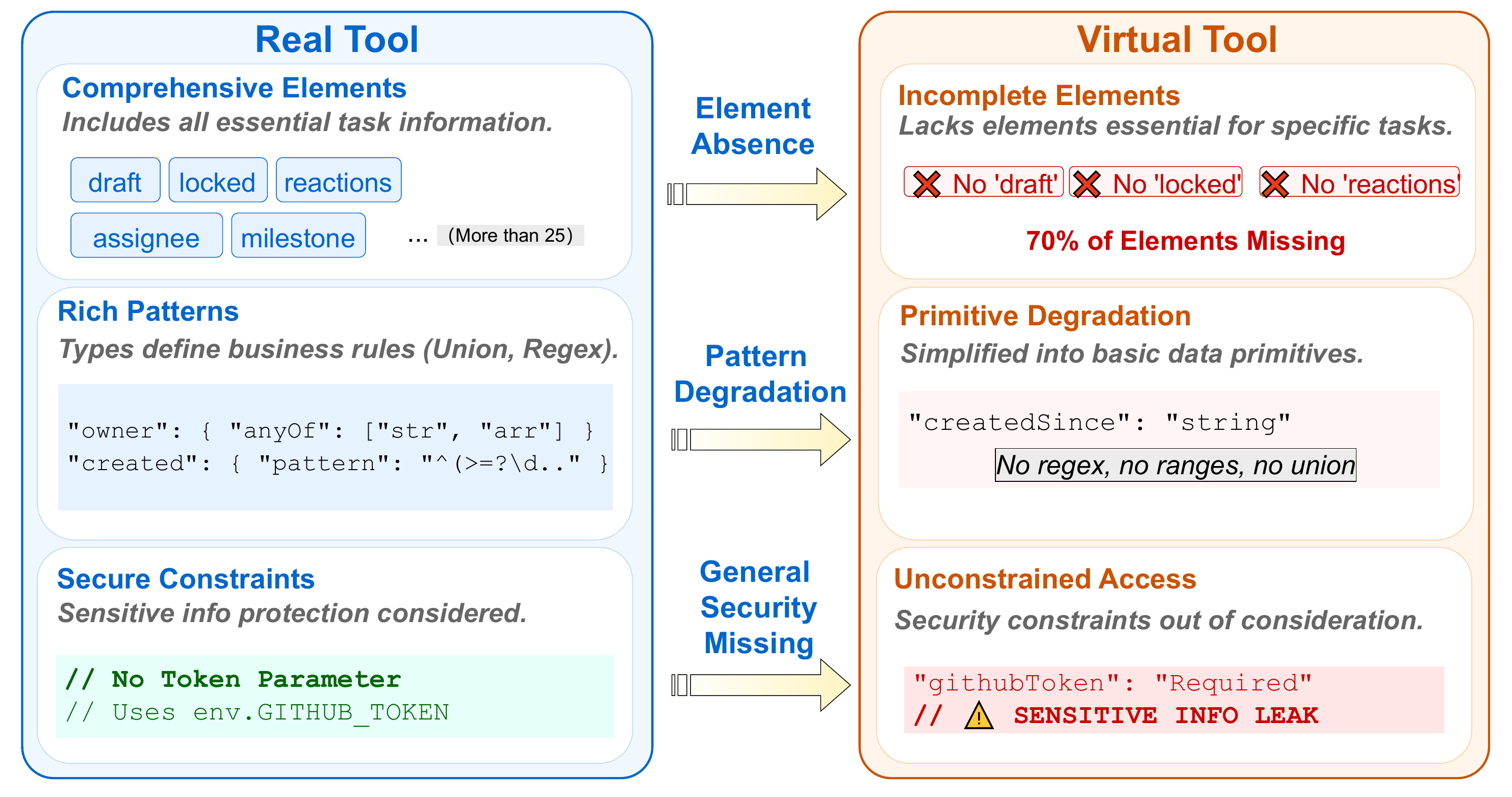}
    \vspace{-4.5ex}
  \caption{The difference between real tools and virtual tools simulated by GPT-4o, exemplified by the \texttt{githubSearchPullRequests} tool (Description: \textit{``Search GitHub pull requests with comprehensive filtering and analysis''}). 
    (1) \textbf{Element Absence}, where virtual tools lack the comprehensive information elements essential for the completion of specific tasks;
    (2) \textbf{Pattern Degradation}, where rich parameter patterns (e.g., unions, regex) are simplified into a data primitive;  
    (3) \textbf{General Security Missing}, where general security constraints, like sensitive information protection, are out of consideration.
    }
    \label{fig:tool_differ_analysis}
    \vspace{-1ex}
\end{figure}

An alternative strategy is to use LLMs as simulators to synthesize tool usage data~\cite{tang2023toolalpacageneralizedtoollearning,li2023apibankcomprehensivebenchmarktoolaugmented,liu2025toolacewinningpointsllm}. 
However, datasets derived from virtual tools are prone to failure in real-world applications, as discrepancies often exist between LLM simulations and real tools deployed in the wild (see Figure~\ref{fig:tool_differ_analysis}).
Furthermore, these datasets focus primarily on task completion, neglecting deviations in tool selection and parameter alignment. As with existing sampling approaches, they ignore negative samples representing intent deviation, a category critical to many learning methods~\cite{yao2022webshop,shen2025shortcutsbenchlargescalerealworldbenchmark}.

In this paper, we propose a method (named {\tool}) to address the intent deviation in tool-using agents.
Focused on the intent-aware tool and parameters, the core design philosophy of {\tool} is twofold.
First, distinct from LLM simulation methods where tool calls are entirely virtual, {\tool} is built upon a \textbf{Real-to-Virtual} strategy.
Beginning with the tool primitives verified in the real environment, {\tool} evolves these real calls into comprehensive virtual trajectories and user requests in a reverse manner.
This potentially narrows the gap with real scenarios while also guaranteeing that critical tool selections can be grounded in user requests.
Second, based on the identification and mutations on the critical parameters, {\tool} generates negatives that simulate diverse intent deviation.
With synthetic data, {\tool} integrates two-stage training to fine-tune backbone LLMs.

Evaluation is performed from three perspectives: data quality, improvement in intent alignment, and generalization ability. 
Regarding data quality, we established eight metrics targeting user requests, tool chains, execution parameters, and response correctness. 
Evaluation results indicate that the data generated by {\tool} maintains high overall quality comparable to other baselines datasets, while outperforming other baselines by an average of 12.62\% in \textit{Tool Relevance}, 11.12\% in \textit{Chain Coherence}, and 11.80\% in \textit{Value Validity}.
For intent alignment, experiments on five mainstream LLMs demonstrate that {\tool} achieves a significant average 35.28\% improvement in $Acc_{\text{task}}$ and 23.27\% in $Acc_{\text{intent}}$, outperforming SOTA baselines by 1.20--42.09\% and 1.17--54.93\% respectively.
Furthermore, even in unseen scenarios, the model enhanced by {\tool} achieves an average 18.22\% and 8.64\% in $Acc_{task}$ and $Acc_{intent}$, compared to the original model across three out-of-distribution datasets.
The contributions are summarized as follows.
\begin{itemize}
    \item 
    We propose {\tool} to address the intent deviation in tool-using agents.
    Combining data synthesis with two-stage training, {\tool} significantly enhances the intent alignment capability.
    \item 
   {\tool} employs a novel data synthesis approach,
   generating virtual trajectories from real calls.
    It can serve as data resources to benchmark intent alignment for agents.
    \item 
    We open-source the package to facilitate reproducibility and future research\footnote{\href{https://anonymous.4open.science/r/TAIAlignment-6C3801/}{https://anonymous.4open.science/r/TAIAlignment-6C3801/}}.
\end{itemize}
\section{Definition of Critical Tool and Parameter}
\label{sec:icp_def}

Given a user request $q$ containing $n$ key elements $\mathcal{K} = \{K_1, K_2, \dots, K_n\}$, where each $K_j$ comprises a specific \textit{\textbf{sub-intention}} and a corresponding \textit{\textbf{key value}} (denoted as $v(K_j)$) to illustrate the sub-intention. 
We assume a mapping where each $K_j$ corresponds to a specific tool call $t_j$ in execution trajectory $\mathcal{T} = \{t_1, t_2, \dots, t_n\}$. For the $j$-th tool $t_j$, let $P(t_j) = \{p_{j1}, p_{j2}, \dots, p_{jm_j}\}$ denote its parameter set, where $v(p_{ji})$ represents the defined value of $p_{ji}$ in the \textit{\textbf{tool primitive}}. We define a parameter $p_{ji} \in P(t_j)$ as an \textit{\textbf{Intent-aware Critical Parameter (ICP)}} if and only if $v(p_{ji})$ is derived from $v(K_j)$, and an \textit{\textbf{Intent-aware Critical Tool (ICT)}} as a tool in $\mathcal{T}$ that contains at least one ICP.
The set of ICPs for the $j$-th tool, denoted as $ICP_j$, and the set of ICTs, denoted as $\mathcal{T}_{ICT}$, are defined as follows:
\begin{equation}
\label{eq:icp_ict}
    \begin{split}
        & ICP_j = \{p_{ji} \in P(t_j) \mid \mathbb{I}_{ICP}(p_{ji}, K_j) = 1\}, \\
        & \mathcal{T}_{ICT} = \{t_j \in \mathcal{T} \mid ICP_j \neq \emptyset\},
    \end{split}
\end{equation}
where $\mathbb{I}_{ICP}(\cdot)$ is an indicator function that determines whether the parameter value is derived from the key value:
\begin{equation}
    \mathbb{I}_{ICP}(p_{ji}, K_j)= 
    \begin{cases}
    1, & \text{if } v(p_{ji}) \text{ is derived from } v(K_j), \\
    0, & \text{otherwise}.
    \end{cases}
\end{equation}

We illustrate these definitions with the user request: ``\textit{Get 1 micro brewery in Denver, CO and recommend 3 same-type breweries.}'' 
In the corresponding tool execution, parameters \texttt{by\_type}, \texttt{by\_city}, \texttt{by\_state}, and \texttt{size} are identified as ICPs because their values in the tool primitive (``micro'', ``Denver'', ``CO'', ``3'') are derived directly from the user's key values. 
Consequently, the tools containing these parameters, specifically \texttt{brewery/search} and \texttt{brewery/random}, are designated as ICTs ($\mathcal{T}_{ICT}$). 
In contrast, the tool \texttt{brewery/by\_id} is excluded from the ICTs, as its parameter \texttt{id} relies solely on internal propagation rather than user's key value.

\section{Methodology}
\label{sec:method}

\begin{figure*}[t]
    \centering
    \includegraphics[width=\linewidth]{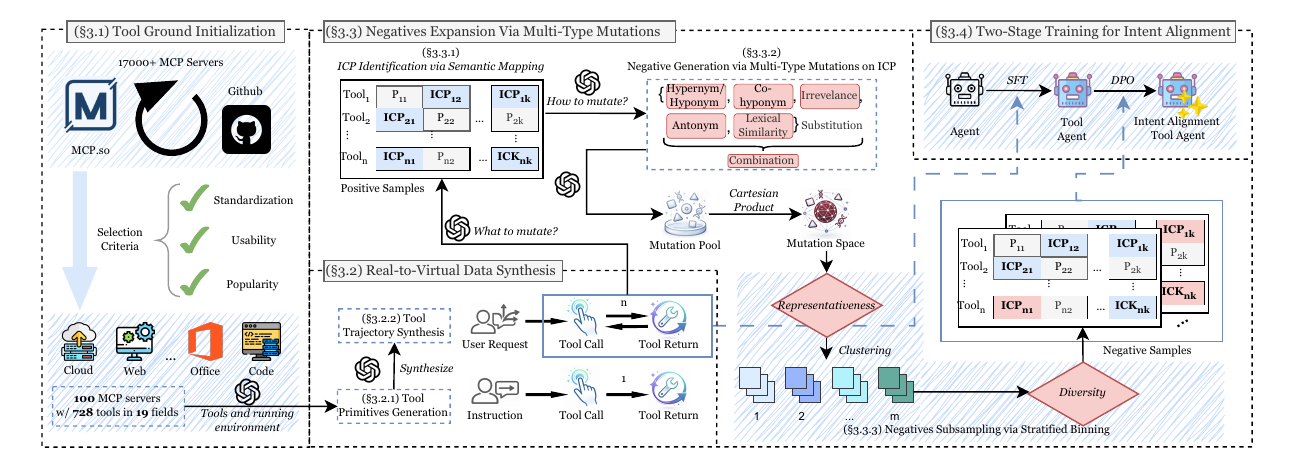} 
    \vspace{-4ex}
    \caption{Overview of the {\tool} for intent-aligned tool-using agents.} 
    \label{fig:method}
    \vspace{-1ex}
\end{figure*}

Figure \ref{fig:method} shows an overview of {\tool}.
First, {\tool} initializes the real tool and the running environment for executability validation.
Second, starting from verified tool primitives (\textit{real aspect}), it reversely synthesizes execution trajectories and user requests (\textit{virtual aspect}). 
Third, through identification and mutation of ICPs, {\tool} augments synthetic data by negatives reflecting multiple intent deviation patterns. 
Fourth, a two-stage training process is integrated to enhance backbone LLMs.
For space limitations, the prompts to drive the LLM in {\tool} are detailed within our package.

\subsection{Tool Ground Initialization}
\label{sec:method-tool_collection}
During this phase, the core is to collect tools and construct a running environment serving as a foundation for the next executability validation.
The following criteria are considered.
\begin{itemize}
    \item 
    \textit{
    \textbf{Standardization}: tool specification and invocation protocol should be standardized.
    }
    \item 
    \textit{
    \textbf{Usability}: tools should exhibit availability, minimizing low-relevance disruptions
    that hinder data synthesis.
    }
    \item 
    \textit{
    \textbf{Popularity}: tools should exhibit popularity, aligning with users’ preference to prioritize mainstream tools when developing agents.
    }
\end{itemize}

In light of the considerations, we choose MCP.so (\url{https://mcp.so/}), the most mainstream distribution platform with over 17,000 MCP servers and numerous tools developed under MCP, as the source for tool collection.
First, we perform stratified sampling on the top 1000 servers of the popularity leaderboard based on domain tags to select representative candidates from major categories.
Then, we follow the inclusion and exclusion criteria (including usage documentation, accessibility, and community recognition) proposed in a previous study~\cite{lin2025largescaleevolvabledatasetmodel} for filtering‌.
After that, we deduplicate‌ the servers based on cosine similarity of BERT embeddings of server names (threshold 0.8) and retain only one with the most starts within the similar ones.
Ultimately, we obtain 100 servers covering 728 tools across 19 fields, and the details are illustrated in our public package.
Meanwhile, we establish a unified MCP client environment to host these servers support subsequent executability validation. 
Under MCP, the client transmits tool call requests with specific parameters. 
A tool calling is deemed valid if the server returns a structured result without protocol-level errors.

\subsection{Real-to-Virtual Data Synthesis}\label{sec:real-to-virtual-data-synthesis}
\subsubsection{Tool Primitive Generation and Executability Validation}
To mitigate the distortion in the tool simulated by LLMs, {\tool} incorporates \textbf{``Real''} consideration‌ into the synthesis process to generate executable tool calling. 
During the process, two steps are performed: tool primitive generation and executability validation.
First, candidate tool primitives are generated by {\tool} using GPT-4o. 
These primitives, comprising tool instructions and their associated usage behaviors, are derived based on the specifications of the collected MCP tools, including functional objectives, parameter definitions, data types, and valid invocation rules. 
To ensure high-quality generation, the prompt includes strict constraints on parameter coverage scenarios (e.g., only required fields or fully populated) and formatting requirements (e.g., unified file paths), guiding the model to generate diverse yet schema-compliant candidates.

Compared to existing simulators based on predefined patterns,
{\tool} anchors the callings in the intrinsic parameter attributes of the tool, ensuring consistency between calling and the tool specification while allowing targeted control over the degree of parameter coverage.

Second, {\tool} verifies the executability of the generated candidates within the constructed running environment.
Candidates are filtered out if they meet any of the following criteria: parameters are invalid and thus fail to invoke an MCP tool effectively; returns contain error or failure messages; or calls exceed the timeout limit.
Finally, the normal calls are treated as real tool primitives for subsequent synthesis.

\subsubsection{Trajectory Synthesis via Multiple Planing Patterns}
\label{sec:tool-chain}
{\tool} combines the verified tool primitive into coherent execution trajectories.
First, {\tool} synthesizes execution trajectories from tool primitives. 
During the step, {\tool} does not arbitrarily arrange tools but integrates the logical patterns relationships between upstream and downstream tools.
Specifically, {\tool} incorporates three distinct patterns, i.e., \textbf{\textit{sequential execution}} for step-by-step subtasks, \textbf{\textit{parallel execution}} for independent subtasks that can run simultaneously, and \textbf{\textit{conditional execution}} where intermediate results are used for subsequent tools. 
Following the patterns, {\tool} generates execution trajectories with the help of GPT-4o.
Notably, the synthesis process utilizes few-shot prompting to control the logical necessity of each step and avoids hallucination by verifying the unique IDs of tool primitives.

\subsubsection{Request Synthesis and Correctness Validation}
Based on verified tool primitives and generated trajectories, {\tool} first generates user requests. 
To ensure alignment between trajectories and user requests, the generation process aims to ensure that both tools and execution parameters are supported by explicit or implicit evidence in the request. 
This alignment is achieved by mapping tool instructions to different sub-intentions in the request, ensuring every action has a corresponding trigger in the natural language input. 
Furthermore, to simulate the real multi-hop user requests, {\tool} referential ambiguity perturbations to obscure technical identifiers while preserving the fundamental semantics of user requests. Specifically, by substituting explicit intermediate parameters (e.g., ``ID 123'') with demonstrative pronouns or natural language references (e.g., ``that order''). Reflecting the realistic gap that user are typically unaware of such technical values and do not include them in their requests. These perturbations force the agent to rely on internal knowledge to deduce the missing links from the context.
After the above steps, {\tool} generates positive samples consisting of user requests paired with their corresponding execution trajectories. 
To further guarantee data reliability, GPT-4o is employed to validate correctness by evaluating consistency between requests and trajectories. 
Samples that successfully pass this verification are annotated with a label indicating the task completion status as the agent response, and they are subsequently utilized as positive samples for training downstream models.

\subsection{Negatives Expansion via Multi-type Mutations}\label{sec:negative-expansion}
Given the positives, {\tool} generates negative samples by mutating ICPs. 
The data generated at this phase are incorporated into the existing synthetic dataset as negative samples.

\subsubsection{ICP Identification via Semantic Mapping}
{\tool} employs a semantic mapping process by guiding the LLM through a structured reasoning workflow. 
First, {\tool} analyzes the user request to identify clearly expressed or implied constraints. 
Second, examines the ground-truth trajectory and gets all parameters value. 
Third, independently verifies the semantic alignment between the parameter values and the identified constraints. Finally, filters out parameters that lack direct evidence in the request. This systematic decomposition and cross-referencing strategy effectively maps ICPs from the tool trajectory.
Formally, for a ground-truth trajectory containing $k$ tool calls, the set of critical parameters for the $j$-th tool call is denoted as $\text{ICP}_j$.

\subsubsection{Negative Generation via Multi-Type Mutations on ICPs}
For each critical parameter $p_{ji} \in \bigcup_{j=1}^k ICP_j$, {\tool} generate a mutation pool $\mathcal{M}(p_{ji})$ containing diverse negative values. 
The construction process is formalized as:
\begin{equation*}
\footnotesize
    \begin{split}
        &\mathcal{M}(p_{ji}) = \{LLM(v_{ji}^*,\mathcal{P} \circ c^{(s)}) \mid c^{(s)} \in \mathcal{C}\} \cup \{v_{ji}^*\} \cup \{\bot\} \\
        &where\ \mathcal{C}=\{c^{(1)},c^{(2)},...,c^{(n)}\}, \bot\ represents\ deletion.
    \end{split}
\end{equation*}
$v_{ji}^*$ is the original correct parameter value. $\mathcal{C}$ represents the set of semantic deviation constraints. $\mathcal{P} \circ c^{(s)}$ denotes the augmented prompt formed by concatenating the base template $\mathcal{P}$ with a specific constraint $c^{(s)}$, guiding the LLM to generate a mutated value. The final pool also includes the original value $\{v_{ji}^*\}$ and a deletion marker $\{\bot\}$.
The core of $\mathcal{C}$ employs six distinct mutations to simulate multiple intent deviations. 
\begin{itemize}
    \item 
    \textit{
    \textbf{Hypernym/Hyponym Substitution}, replacing an entity with its superordinate or subordinate concept (e.g., from ``iPhone'' to ``phone'').
    }
    \item 
    \textit{
    \textbf{Co-hyponym Substitution}, substituting an entity with a different instance from the same semantic concept (e.g., from ``flight'' to ``train''). 
    }
    \item 
    \textit{
    \textbf{Irrelevance Substitution}, replacing a concept with a completely unrelated term to break semantic relevance (e.g., from ``smartphone'' to ``banana'').
    }
    \item 
    \textit{
    \textbf{Antonym Substitution}, substituting a word with its antonym (e.g., from ``cheapest'' to ``most expensive'').
    }
    \item 
    \textit{
    \textbf{Lexical Similarity Substitution}, substituting a lexically similar but semantically distinct term (e.g., from  ``audiobook'' to ``book review'').
    }
    \item 
    \textit{
    \textbf{Combination}, applying a composite strategy that integrates any combination of the aforementioned mutation rules to generate complex perturbations.
    }
\end{itemize}

Guided by the mutations, {\tool} utilizes GPT-4o with a carefully designed prompt $\mathcal{P}$ to generate a batch of unique values covering the first five substitution. 
Crucially, to ensure that the negative values in $\mathcal{M}(p_{ji})$ are not trivial specification errors easily caught by parsing, but rather genuine semantic logic errors, the prompt explicitly requires the LLM to maintain the same part of speech and data type as the original values. 
For \textit{\textbf{Combination}}, {\tool} constructs a comprehensive mutation space via the Cartesian~\cite{weisstein2025cartesian} of individual mutations. 

To ensure the quality of these negative samples, mutation complexity is quantified using a weighted score that integrates the mutation ratio. 
Specifically, deviations for numerical values are calculated via relative numerical difference, while those for text values are assessed using a BERT-based semantic distance. 
Any combinations with a complexity score falling below a threshold $\theta$ are filtered out.

\subsubsection{Negatives Subsampling via Stratified Binning}

To avoid the redundancy and imbalance, {\tool} samples a representative subset that preserves the diversity of the original ones.
Two core principles are considered:
    (1) \textbf{Representativeness}, mutations with different structural characteristics should be proportionally represented, avoiding the neglect of minority mutation structures;
    and (2) \textbf{Diversity}, within mutations of the same structural type, the sampled combinations should cover the full range of mutation significance, ensuring that both mild and severe mutations are included.

To begin, {\tool} utilizes mask clustering to group mutations into clusters $\{C_g\}$ according to structural patterns, defined by a binary matrix $\mathbf{M} \in \{0,1\}^{k \times M}$ where 1 indicates a mutation site.
Following this, {\tool} applies an adaptive quota allocation strategy. Given a total target budget $N$, the quota $n_g$ for cluster $C_g$ is calculated as:
\begin{equation}
    n_g = \max\left(1, \left\lfloor N \cdot \frac{|C_g|}{|\mathcal{S}|} \right\rfloor\right) + \delta_g,
\end{equation}
where $\delta_g$ is an adjustment term derived from the remaining budget to ensure $\sum_{g=1}^G n_g = N$. This method ensures the final sample size target is met, striking a balance between proportionality and fairness to avoid neglecting rare patterns.
Finally, each cluster $C_g$ is sorted by mutation score and divided into equal-sized bins $\{B_{g,l}\}$. 
By evenly distributing quotas and randomly sampling, {\tool} employs a stratified binning strategy that effectively mitigates bias toward specific mutation intensities to construct the final dataset $\mathcal{S}_{\text{final}}$:
\begin{equation}
    \mathcal{S}_{\text{final}} = \bigcup_{g=1}^G \bigcup_{l=1}^L \text{Sample}\left(B_{g,l}, \frac{n_g}{L}\right).
\end{equation}

\subsection{Two-Stage Training for Intent Alignment}
\label{sec:method-training}

{\tool} employs a two-stage training paradigm that combines SFT~\cite{ouyang2022traininglanguagemodelsfollow} for basic capability calibration and DPO~\cite{rafailov2024directpreferenceoptimizationlanguage} for fine-grained intent alignment. 
The primary objective of this two-stage is to ensure that the LLM not only acquires fundamental tool-using knowledge but also aligns its behaviors precisely with user intent. 

In the first stage, SFT is performed using positive samples to calibrate the basic tool-using capability. This step is crucial because it prevents LLM from generating random or invalid calls. 
The second stage employs DPO to address the limitations of relying solely on positive samples, as syntactically valid tool behaviors may still deviate from user intent.
Specifically,
{\tool} pairs the original \( c_w \) and mutated negative samples \( c_l \). 
These pairs form the DPO preference dataset \( \mathcal{D}_{\text{DPO}} = \{(x, c_w, c_l)\} \), where \( x \) denotes the comprehensive input including the user requests, and candidate tool set.
DPO is to optimize LLMs by guiding it to generate the desired trajectory \( c_w \) while avoiding \( c_l \). 
The loss function is defined as:
\[
\resizebox{0.5\textwidth}{!}{$\mathcal{L}_{\text{DPO}} = -\mathbb{E}_{\mathcal{D}_{\text{DPO}}} \log\sigma\left( \beta \left[ \log\frac{\pi_\theta(c_w|x)}{\pi_{\text{ref}}(c_w|x)} - \log\frac{\pi_\theta(c_l|x)}{\pi_{\text{ref}}(c_l|x)} \right] \right)$}
\]
where \( \sigma(\cdot) \) is the logistic function, \( \beta \) is a weighting parameter controlling the deviation of the policy model \( \pi_\theta \) (the model to be optimized) from the reference model \( \pi_{\text{ref}} \) (the SFT-stage model), and \( \mathbb{E}_{\mathcal{D}_{\text{DPO}}} \) denotes the expectation over the DPO preference dataset.

\section{Experimental Design}

\subsection{Research Questions}
Three formulated research questions focus on the quality of synthetic data and its benefits for two-stage LLM training. 

\textit{\textbf{RQ1 (Data Quality): Can {\tool} generate high-quality synthetic data?} 
We quantify the quality from the synthetic trajectory, as well as the user requests.}

\textit{\textbf{RQ2 (Intent Alignment): Can {\tool} effectively mitigate the intention deviation in tool-using agents?} 
We validate whether {\tool} helps the backbone LLMs address intent deviation while maintaining usability within our synthetic data.} 

\textit{\textbf{RQ3 (Generalization): Can {\tool} help LLM's generalize to unseen tool-using scenarios?}} 
We extend beyond our synthetic data to investigate whether the LLMs enhanced by {\tool} exhibit superior performance in unseen scenarios.

\subsection{Experiment for Data Quality (RQ1)}
\label{sec:experiment_rq1}
For RQ1, we compare with the data sampling from five prevalent datasets for tool-using agents (ToolBench~\cite{qin2023toolllmfacilitatinglargelanguage},ToolAlpaca~\cite{tang2023toolalpacageneralizedtoollearning}, STE ~\cite{wang2024llmsimaginariumtoollearning}, GPT4Tools~\cite{yang2023gpt4toolsteachinglargelanguage} and PEToolBench~\cite{xu2025petoolllmpersonalizedtoollearning}), employing both human and LLM-as-a-judge~\cite{gu2025surveyllmasajudge} to assess the quality of synthetic data from the following dimensions.

\begin{itemize}
  \item 
    \textit{
    \textbf{User Request}, where we evaluate the Linguistic Naturalness of the user requests, focusing on grammaticality and fluency following standard NLG evaluation criteria~\cite{gatt2018surveystateartnatural}.
    }
    
    \item
    \textit{
    \textbf{Tool Chain}, where we assess the appropriateness of tool selection and the logical coherence of the execution sequence. 
    It includes three sub-dimensions: \textbf{Tool Relevance} (the relevance of the selected tools to the user intent in the request), \textbf{Call Conciseness} (the proportion of non-redundant calls in the tool chain), and \textbf{Chain Coherence} (the logical coherence of the tool execution sequence annotated by human raters).
    }

    \item 
    \textit{
    \textbf{Execution Parameter},
    where we measure the executability and consistency of the execution parameters during the tool calling phase. It includes three sub-dimensions: \textbf{Name Validity} (the percentage of the consistent parameter names with those in tool specifications), \textbf{Value Validity} (the compliance with specified data types and ranges), and \textbf{Value Consistency} (the alignment between critical parameter values to the information in user request).
}
    
    \item 
    \textit{
    \textbf{Response Correctness}:
    where we calculate the percentage of correct agent responses, determining whether the final agent response correctly reflects task completion or not based on user request and execution trajectory.
    }
\end{itemize}
 
Except for \textbf{\textit{Call Conciseness}}, which can be calculated by exact string matching, the remaining seven sub-dimensions are evaluated by GPT-4o on a unified 5-point Likert scale~\cite{encyclopedia5010018}. We establish strict criteria for this scale: a score of \textbf{1.0} indicates critical defects (e.g., hallucinations, schema violations, or complete irrelevance), whereas a score of \textbf{5.0} represents optimal quality (e.g., fully executable, logically coherent, and linguistically fluent). 
Detailed scoring rubrics and prompts for LLM-as-a-judge are available in our package.

\subsection{Experiment for Intent Alignment (RQ2)}
\label{sec:experiment_rq2}
For RQ2, 
we use Llama-3.1-8B-Instruct, Qwen3-8B, Mistral-7B, Qwen2.5-7B-Instruct, and Deepseek-llm-7b-chat as subjects to investigate whether {\tool} can enhance them with two-stage training and synthetic data.
For measurement, five metrics are used, including the first two established in the literature (donated as $\bigstar$) whereas the latter three are newly proposed metrics (donated as $\Delta$).
\begin{itemize}
    \item 
    \textit{
    $\bigstar Acc_{task}$: percentage of completed tasks to the total, excluding the cases of task failures.
    }
    \item 
    \textit{
    $\bigstar Acc_{calling}$: percentage of valid tool calling, focusing on the executability of tool calling.
    }
    \item 
    \textit{
    $\Delta Pre_{tool}$: percentage of correct tool selection (relevant to user request) in the tool chain, excluding unnecessary and redundant cases. 
    }
    \item 
    \textit{
    $\Delta Rec_{tool}$: percentage of correctly selected tools to all required tools, assessing the completeness of tool selection for task completion. 
    }
    \item 
    \textit{
    $\Delta Acc_{ICP}$: percentage of ICPs consistent with user requests, restricted to cases of correct tool selection.
    }
    \item 
    \textit{
    $\Delta ACC_{intent}$: percentage of the cases where the agent's actions faithfully adhere to the user's original intent, to compensate for the limitation that $Acc_{task}$ merely reflects task success or failure without capturing whether the agent's behaviors align with user intent. 
    }
\end{itemize}

For comparison, we also introduce three open-source tool-using LLMs and one advanced closed-source LLM as baselines, i.e., ToolLLaMA-2-7b-v2~\cite{qin2023toolllmfacilitatinglargelanguage}, Lynx-7B~\cite{li2023apibankcomprehensivebenchmarktoolaugmented}, ToolAlpaca-7B~\cite{tang2023toolalpacageneralizedtoollearning}) and GPT-5 nano~\cite{openai2025gpt5nano}.
For $Acc_{task}$ and $Acc_{calling}$ we follow the evaluation method in the previous study~\cite{qin2023toolllmfacilitatinglargelanguage,wang2024gtabenchmarkgeneraltool}. 
For $\Delta Pre_{tool}$ and $\Delta Rec_{tool}$, which are determined by comparing the intersection of predicted and gold tool selection against the total predicted/required tools, $\Delta Acc_{ICP}$ is validated by exact matching (lowercase, space-stripped) between predicted and gold ICP values. As for $\Delta ACC_{intent}$, it is further determined by incorporating the tool selection trajectory.

\subsection{Experiment for Generalization (RQ3)}
\label{sec:experiment_rq3}
For RQ3, we evaluate the generalization of {\tool} on three distinct out-of-distribution (OOD) datasets: \textbf{ToolBench}~\cite{qin2023toolllmfacilitatinglargelanguage}, \textbf{DICE-Bench}~\cite{jang2025dicebenchevaluatingtoolusecapabilities}, and \textbf{ACEBench}~\cite{chen2025acebenchwinsmatchpoint}.
These datasets represent unseen domains and varying levels of complexity, spanning from long-chain reasoning with massive tools in \textbf{ToolBench}, to context integration amidst fragmented information in multi-party dialogues in \textbf{DICE-Bench}, and finally to the handling of imperfect instructions in \textbf{ACEBench}.

Considering the annotation of these OOD datasets, our evaluation focuses solely on task completion, rather than examining the agent's internal execution trajectory.
Two metrics are reused to assess the performance of task completion in these unseen tasks, i.e, $\bigstar Acc_{task}$ and $\Delta ACC_{intent}$. 
The evaluation methods are the same as those in Section \ref{sec:experiment_rq3}.

\vspace{-1ex}

\section{Results}

\subsection{Results of Data Quality (RQ1)}
\begin{figure}[!t]
    \centering
    \includegraphics[width=0.8\linewidth]{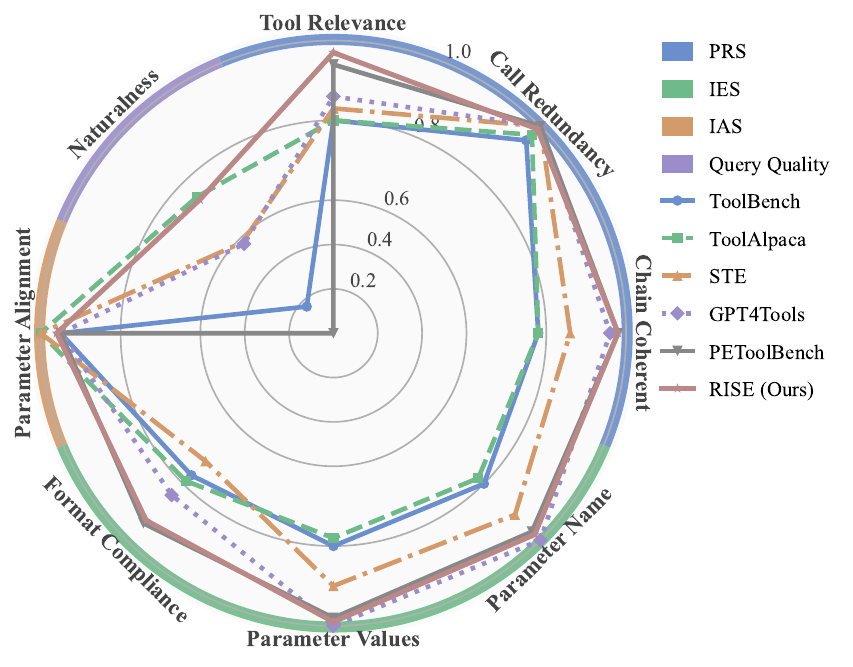}
    \vspace{-2.5ex}
    \caption{Results of quality evaluation of the synthetic data}
    \vspace{-1ex}
    \label{fig:dataset_quality_dimensions}
\end{figure}

Figure \ref{fig:dataset_quality_dimensions} shows the results of the data quality based on LLM-as-a-judge.
In general, {\tool} demonstrates superior or highly competitive performance across all dimensions.
Next, we elaborate on the key findings derived from the results and present the supporting evidence.
First, \textbf{Synthesized requests and responses achieve effectiveness comparable to existing datasets. }
For \textit{Naturalness}, {\tool} achieves a score of 0.96, which is on par with established baselines like ToolBench (0.95) and GPT4Tools (0.95), ensuring the fluency of user requests. 
Crucially, for \textit{Response Correctness}, {\tool}  scores 0.74, significantly outperforming ToolBench (0.17), STE (0.58), and GPT4Tools (0.57).

Second, \textbf{Trajectories generated by {\tool} outperform the baseline in internal rationality and user intent alignment.} 
For internal rationality, {\tool} ties for the top score in \textit{Chain Coherence} (0.98) and achieved near-perfect scores for \textit{Value Validity} (0.99) and \textit{Name Validity} (0.98). 
For intent alignment, {\tool} reaches the highest performance in \textit{Tool Relevance} (0.97), and \textit{Value Consistency} (0.93), significantly outperforming ToolBench (0.80/0.77) and ToolAlpaca (0.80/0.79). highlighting its advantage in intent alignment.

\subsection{Results of Intent Alignment (RQ2)}
We randomly partitioned the synthesized dataset into a training set and a test set with a ratio of 8:2. Table \ref{tab:main_results} presents the comparative performance of 5 representative backbone models on the test set.
The rows correspond to different training configurations: 
\textit{Raw} denotes the off-the-shelf base LLMs without any fine-tuning; 
\textit{+SFT} refers to models supervised fine-tuned using standard positive samples; 
\textit{+DPO} indicates models trained directly on positive-negative pairs, without prior supervised fine-tuning;
and \textit{Ours} represents the final models trained using our complete two-stage paradigm.
The key findings are elaborated below.

\textbf{{\tool} significantly reduces intention deviation, ensuring that LLMs adhere to user intents.}
{\tool} demonstrates robust performance improvements across all baselines. Specifically, it achieves an \textbf{average improvement of 35.28\%} in $\Delta Acc_{\text{Intent}}$ and \textbf{23.27\%} in $\Delta Acc_{\text{ICP}}$ compared to Raw models.
This substantial gain is largely attributed to the explicit modeling of Intent-Critical Parameters (ICPs). Unlike static ``required parameters'' defined by tool schemas, ICPs are derived dynamically from task-specific constraints. By incorporating these constraints, {\tool} effectively bridges the gap between user intent and execution, as evidenced by the consistent $\Delta Acc_{\text{ICP}}$ gains over SFT baselines (e.g., Llama-3.1 rises from 60.02\% to 62.19\%).

\textbf{{\tool} does not compromise general tool-using performance at the expense of intent alignment; instead, it improves it.}
On average, {\tool} boosts the task completion rate ($\bigstar Acc_{\text{task}}$) by \textbf{27.28\%} and elevates tool selection metrics ($\Delta \text{Precision}_{\text{tool}}$ and $\Delta \text{Recall}_{\text{tool}}$) by \textbf{27.14\%} over Raw baselines.
In addition, the results highlight the necessity of two-stage training. 
While SFT establishes foundational capabilities, skipping it (i.e., the \textit{+DPO} setting) leads to suboptimal performance (e.g., Qwen2.5-7B drops from 62.73\% to 61.92\% in $\bigstar Acc_{\text{calling}}$). 

\subsection{Results of Generalization (RQ3)}
\begin{figure}[!t]
    \centering
    \includegraphics[width=0.5\textwidth]{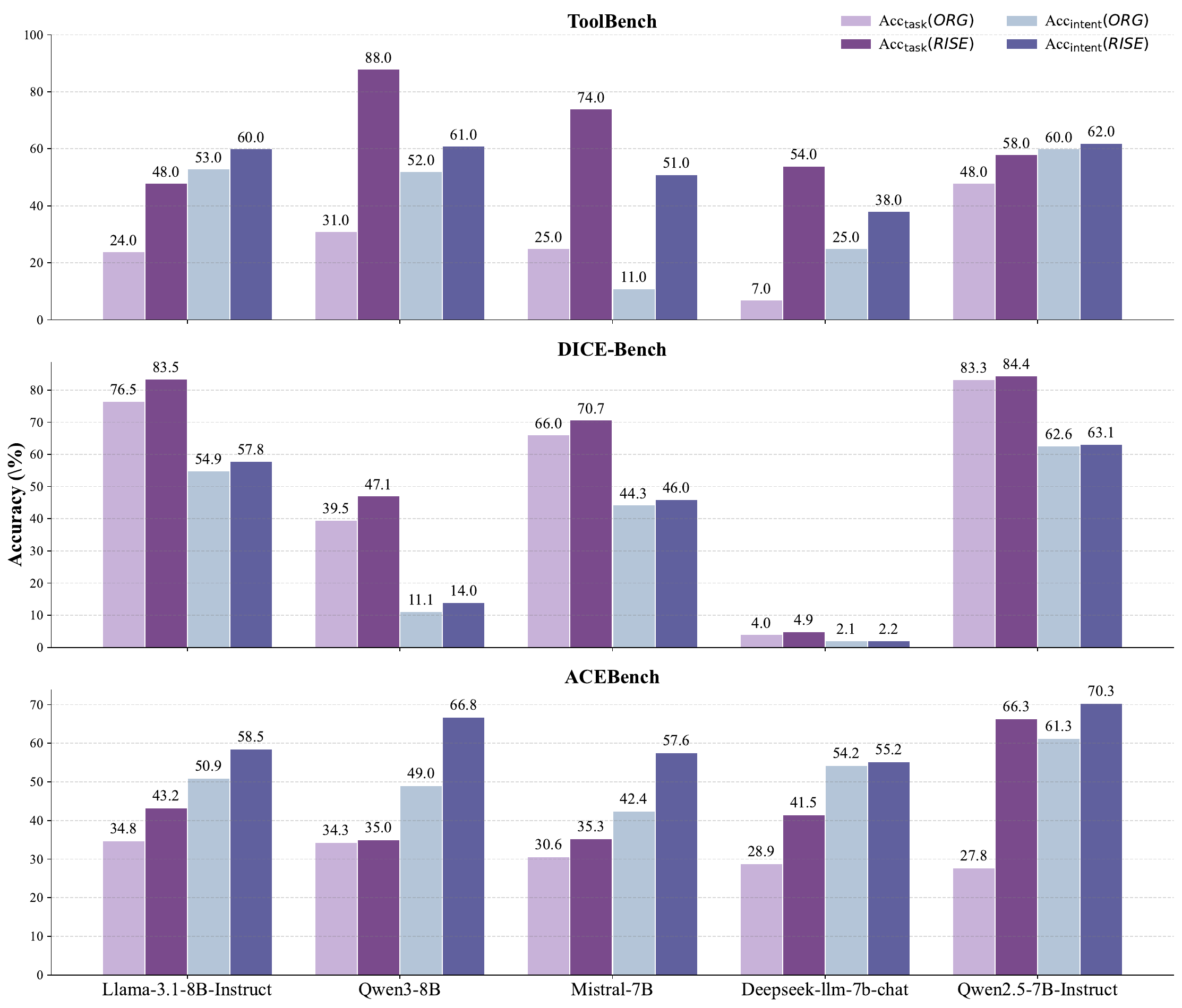}
    \vspace{-4ex}
    \caption{
    Performance Comparison of the original model (ORG) and enhanced by {\tool} (Ours) on OOD datasets}
    \label{fig:rq3_performance}
    \vspace{-1ex}
\end{figure}

Figure \ref{fig:rq3_performance} illustrates the performance comparison between original LLMs and those enhanced by {\tool} on three OOD datasets.
In summary, the enhanced LLMs demonstrate superior performance even in unseen scenarios.
Integrating the characteristics of the three datasets, we obtained the following findings.
First, \textbf{{\tool} dramatically improves reasoning capabilities in complex, long-horizon tasks.}
On \textbf{ToolBench}, which requires handling massive tools and long execution chains, LLMs enhanced by {\tool} exhibit substantial performance leaps. 
{\tool} reaches an average 37.4\% and 14.2\% imprvement on $Acc_{task}$, $Acc_{intent}$. Specifically, Qwen3-8B shows the most significant gain, with $Acc_{task}$ from 31\% (ORG) to 88\% ({\tool}) and Mistral-7B also realizes a dramatic improvement, with $Acc_{task}$ jumping from 25\% to 74\% and $Acc_{intent}$ rising from a mere 11\% to 51\%.  
Even the relatively low-performing deepseek-llm-7b-chat sees its $Acc_{task}$ improve from 7\% to 54\%. 

Second, \textbf{{\tool} demonstrates strong adaptability in lengthy and ambiguous user requests.}
In scenarios involving ambiguous user inputs (ACEBench) and multi-party dialogue context (DICE-Bench), {\tool} exhibits superior generalization capabilities. 
For ACEBench, an average 13.0\% and 10.1\% improvement on $Acc_{task}$, $Acc_{intent}$ and Qwen2.5-7B-Instruct achieves the most notable enhancement, with $Acc_{task}$ surging from 27.75\% (ORG) to 66.30\% ({\tool}) and $Acc_{intent}$ increasing from 61.30\% to 70.30\%. On DICE-Bench, a 4.2\% average improvement of $Acc_{task}$ and 1.6\% on $Acc_{intent}$ (note that deepseek-llm-7b-chat exhibits lower absolute performance on DICE-Bench due to its context length limit). 
These results confirm that {\tool} enables LLMs to master underlying tool-using logic, ensuring robust generalization despite domain shifts.

\begin{table}[t]
\centering
\caption{Performances of the trained LLMs on the synthetic data}
\vspace{-2ex}
\label{tab:main_results}
\resizebox{\linewidth}{!}{%
\begin{tabular}{lcccccc} %
\toprule
\textbf{Method} & $\bigstar \textbf{Acc}_{\textbf{task}}$ & $\bigstar \textbf{Acc}_{\textbf{calling}}$ & $\Delta \textbf{Acc}_{\textbf{ICP}}$ & $\Delta \textbf{Acc}_{\textbf{Intent}}$ & $\Delta \textbf{Precision}_{\textbf{tool}}$ & $\Delta \textbf{Recall}_{\textbf{tool}}$ \\
\midrule
\multicolumn{7}{c}{\textbf{Llama-3.1-8B-Instruct}} \\
\midrule
Raw & 65.60\% & 62.09\% & 32.38\% & 8.33\% & 74.82\% & 48.34\% \\
+SFT & 88.58\% & 98.33\% & 60.02\% & 47.22\% & 83.20\% & 79.94\% \\
+DPO & 68.83\% & 63.08\% & 38.64\% & 8.33\% & 73.22\% & 47.99\% \\
\textbf{Ours} & \textbf{91.98\%} & \textbf{98.81\%} & \textbf{62.19\%} & \textbf{50.15\%} & \textbf{85.44\%} & \textbf{82.66\%} \\
\midrule
\multicolumn{7}{c}{\textbf{Qwen2.5-7B-Instruct}} \\
\midrule
Raw & 60.03\% & 62.73\% & 48.33\% & 41.67\% & 64.10\% & 65.73\% \\
+SFT & 92.28\% & 97.70\% & 55.98\% & 67.90\% & 85.73\% & 80.16\% \\
+DPO & 58.49\% & 61.92\% & 49.68\% & 39.66\% & 62.73\% & 63.94\% \\
\textbf{Ours} & \textbf{92.29\%} & \textbf{97.84\%} & \textbf{58.24\%} & \textbf{69.13\%} & \textbf{86.83\%} & \textbf{81.63\%} \\
\midrule
\multicolumn{7}{c}{\textbf{deepseek-llm-7b-chat}} \\
\midrule
Raw & 47.84\% & 24.65\% & 17.45\% & 5.86\% & 18.08\% & 21.81\% \\
+SFT & 89.81\% & 81.28\% & 48.43\% & 60.34\% & 77.39\% & 74.66\% \\
+DPO & 50.93\% & 21.44\% & 20.85\% & 7.56\% & 13.12\% & 15.10\% \\
\textbf{Ours} & \textbf{89.93\%} & \textbf{81.24\%} & \textbf{50.86\%} & \textbf{60.79\%} & \textbf{79.36\%} & \textbf{75.81\%} \\
\midrule
\multicolumn{7}{c}{\textbf{Qwen3-8B}} \\
\midrule
Raw & 67.59\% & 60.77\% & 32.38\% & 8.33\% & 74.82\% & 48.34\% \\
+SFT & 88.58\% & 83.68\% & 58.88\% & 44.14\% & 83.07\% & 84.16\% \\
+DPO & 65.90\% & 63.34\% & 47.06\% & 24.38\% & 56.80\% & 58.78\% \\
\textbf{Ours} & \textbf{90.90\%} & \textbf{95.11\%} & \textbf{60.62\%} & \textbf{44.60\%} & \textbf{84.26\%} & \textbf{84.23\%} \\
\midrule
\multicolumn{7}{c}{\textbf{Mistral-7B}} \\
\midrule
Raw & 70.83\% & 71.32\% & 32.10\% & 11.57\% & 60.91\% & 53.57\% \\
+SFT & 83.03\% & 72.90\% & 45.77\% & 26.70\% & 67.02\% & 71.52\% \\
+DPO & 69.75\% & 69.90\% & 35.84\% & 11.76\% & 60.30\% & 53.51\% \\
\textbf{Ours} & \textbf{83.18\%} & \textbf{73.41\%} & \textbf{47.06\%} & \textbf{27.47\%} & \textbf{68.56\%} & \textbf{73.17\%} \\
\midrule
\multicolumn{7}{c}{\textbf{Baselines}} \\
\midrule
TL2 & \textbf{87.38\%} & 72.31\% & 32.68\% & \textbf{38.21\%} & 53.27\% & 56.42\% \\
LYN & 6.35\% & \textbf{94.26\%} & 1.75\% & 3.51\% & 22.36\% & 26.90\% \\
TAL & 41.27\% & 63.94\% & 9.63\% & 17.18\% & 51.30\% & 46.30\% \\
G5N & 68.99\% & 92.94\% &\textbf{ 57.13\%} & 29.61\% & \textbf{59.39\%} & \textbf{79.40\%}\\
\bottomrule
\end{tabular}%
}
{\tiny Legends: ToolLLaMA-2-7b-v2 (TL2), Lynx-7b (LYN), ToolAlpaca-7B (TAL), GPT-5-nano (G5N)}
\end{table}

\vspace{-0.5em}

\section{Related Work}

\paragraph{Benchmarking for Tool-using Agents}
Recent work has proposed diverse benchmarks to evaluate tool-using agents. 
One line of research constructs large-scale benchmarks by simulating tool call and feedback with LLMs~\cite{liu2025toolacewinningpointsllm}, which scales well but relies on virtual tools and thus lacks executability. 
Another line samples trajectories from real interaction environments~\cite{wang2024llmsimaginariumtoollearning,yao2024taubenchbenchmarktoolagentuserinteraction}, ensuring authenticity at the cost of high collection overhead and limited control over trajectory diversity. 
Moreover, existing benchmarks mainly emphasize tool selection~\cite{huang2024metatoolbenchmarklargelanguage,ning2024wtuevalwhetherornottoolusage} or task completion accuracy~\cite{zhuang2023toolqadatasetllmquestion}, while largely overlooking subtile issues during execution that can induce intent deviation~\cite{zheng2025all,ge2025can}.

\paragraph{Methods for Tool-Using Augmentation} Methods for enhancing tool-using agents include in-context learning~\cite{Paranjape2023ARTAM,yuan2024easytoolenhancingllmbasedagents}, supervised fine-tuning~\cite{Hsieh2023ToolDE}, and post-training~\cite{rafailov2024directpreferenceoptimizationlanguage,bi2025rewardguidancerubricspromoting}. All these approaches rely heavily on high-quality training data, obtained either through manual construction or automated synthesis. 
Early work collected large-scale execution trajectories from strong closed-source LLMs with manual annotation~\cite{qin2023toolllmfacilitatinglargelanguage}, which is costly and limited in diversity. 
More recent studies adopt prompt-guided LLMs to synthesize training data~\cite{jung2025diatool,xu2025petoolllmpersonalizedtoollearning,chen2025advancingtoolaugmentedlargelanguage,mei2025a1steeptesttimescaling}, improving scalability but often sacrificing authenticity and generalization. 
Both paradigms mainly focus on the task completion, leaving intent deviation unaddressed.

\vspace{-0.5em}

\section{Conclusion}
In this paper, we introduce {\tool}, a method to address intent deviation in tool-using agents. 
Guided by ``Real-to-Virtual'' principle, {\tool} synthesizes virtual trajectories derived from real tool calls and diverse negative samples generated via multi-type mutations on ICPs.
With synthetic data, {\tool} enhances LLMs through a two-stage training. 
This work advances intent alignment in tool-using agents through novel data synthesis and training strategies, while providing valuable benchmarks for the community.

\bibliographystyle{named}
\bibliography{ijcai26}

\end{document}